\documentclass[runningheads]{llncs}
\usepackage{cite}
\usepackage{amsmath,amssymb,amsfonts}
\usepackage{mathtools}
\usepackage{epsfig}
\usepackage[mathscr]{euscript}
\usepackage{color}
\usepackage{booktabs} 
\usepackage{stmaryrd} 
\usepackage{bbm} 
\usepackage{blindtext, comment}
\usepackage{rotating} 
\usepackage{tikz}
\usepackage{standalone}
\usetikzlibrary{arrows, patterns}
\usetikzlibrary{decorations.pathreplacing,angles,quotes}
\usepackage{framed, standalone}
\usepackage{graphicx}
\usepackage{capt-of}
\usepackage{balance}
\usepackage{makecell} 
\usepackage{afterpage}
\usepackage[linesnumbered,lined,boxed,commentsnumbered,ruled,vlined]{algorithm2e}
\usepackage{setspace}
\usepackage{textcomp}
\usepackage{xcolor}
\def\BibTeX{{\rm B\kern-.05em{\sc i\kern-.025em b}\kern-.08em
    T\kern-.1667em\lower.7ex\hbox{E}\kern-.125emX}}
\begin{document}
\title{
MQLV: Optimal Policy of Money Management in Retail Banking with Q-Learning
}
\titlerunning{Q-Learning of Money Management in Retail Banking}
%
\author{Jeremy Charlier \inst{1,2}
\and
Gaston Ormazabal \inst{2}
\and
Radu State \inst{1}
\and
Jean Hilger\inst{3}}

\authorrunning{J. Charlier et al.}
\institute{University of Luxembourg, L-1855 Luxembourg, Luxembourg  \\
\email{\{name.surname@\}@uni.lu}
\and
Columbia University, New York NY 10027, USA  \\
\email{\{jjc2292,gso7@\}@columbia.edu}
\and
BCEE, L-1160 Luxembourg, Luxembourg  \\
\email{j.hilger@bcee.lu}\\
}
\maketitle

\begin{abstract}
Reinforcement learning has become one of the best approach to train a computer game emulator capable of human level performance. In a reinforcement learning approach, an optimal value function is learned across a set of actions, or decisions, that leads to a set of states giving different rewards, with the objective to maximize the overall reward. A policy assigns to each state-action pairs an expected return. We call an optimal policy a policy for which the value function is optimal. QLBS, Q-Learner in the Black-Scholes(-Merton) Worlds, applies the reinforcement learning concepts, and noticeably, the popular Q-learning algorithm, to the financial stochastic model of Black, Scholes and Merton. 
It is, however, specifically optimized for the geometric Brownian motion and the vanilla options. Its range of application is, therefore, limited to vanilla option pricing within financial markets. 
We propose MQLV, Modified Q-Learner for the Vasicek model, a new reinforcement learning approach that determines the optimal policy of money management based on the aggregated financial transactions of the clients. It unlocks new frontiers to establish personalized credit card limits or to fulfill bank loan applications, targeting the retail banking industry. MQLV extends the simulation to mean reverting stochastic diffusion processes and 
it uses a digital function, a Heaviside step function expressed in its discrete form, to estimate the probability of a future event such as a payment default. In our experiments, we first show the similarities between a set of historical financial transactions and Vasicek generated transactions and, then, we underline the potential of MQLV on generated Monte Carlo simulations. Finally, MQLV is the first Q-learning Vasicek-based 
methodology 
addressing transparent decision making processes in retail banking. 
\end{abstract}
\section{Introduction} \label{sec::intro}
A major goal of the reinforcement learning (RL) and Machine Learning (ML) community is to build efficient representations of the current environment to solve complex tasks. In RL, an agent relies on multiple sensory inputs and past experience to derive a set of plausible actions to solve a new situation \cite{mnih2013playing}. While the initial idea around RL is not new \cite{sutton1984temporal,watkins1989learning,williams1987class}, significant progress has been achieved recently by combining neural networks and Deep Learning (DL) with RL. The progress of DL \cite{krizhevsky2012imagenet,sermanet2013pedestrian} has allowed the development of a novel agent combining RL with a class of deep artificial neural networks \cite{mnih2013playing,mnih2015human} resulting in Deep Q Network (DQN). The Q refers to the Q-learning algorithm introduced in \cite{watkins1992q}. It is an incremental method that successively improves its evaluations of the quality of the state-action pairs. The DQN approach achieves human level performance on Atari video games using unprocessed pixels as inputs. In \cite{van2016deep}, deep RL with double Q-Learning was proposed to challenge the DQN approach while trying to reduce the overestimation of the action values, a well-known drawback of the Q-learning and DQN methodologies. 
The extension of the DQN approach from discrete 
to continuous action domain, directly from the raw pixels to inputs, was successfully achieved for various simulated tasks \cite{lillicrap2015continuous}.  \\

Nonetheless, most of the proposed models focused on gaming theory and computer game simulation and very few to the financial world. 
In QLBS \cite{halperin2017qlbs}, a RL approach is applied to the Black, Scholes and Merton financial framework for derivatives \cite{black1973pricing,merton1973theory}, a cornerstone of the modern quantitative finance. In the BSM model, the dynamic of a stock market is defined as following a Geometric Brownian Motion (GBM) to estimate the price of a vanilla option on a stock \cite{wilmott2013paul}. A vanilla option is an option that gives the holder the right to buy or sell the underlying asset, a stock, at maturity for a certain price, the strike price. QLBS is one of the first approach to propose a complete RL framework 
for finance. As mentioned by the author, a certain number of topics are, however, not covered in the approach. For instance, it is specifically designed for vanilla options and it fails to address any other type of financial 
applications. Additionally, the initial generated paths rely on the popular GBM but there exist a significant number of other popular stochastic models depending on the market dynamics \cite{hull2003options}.  \\

In this work, we describe a RL approach tailored for personal recommendation in retail banking regarding money management to be used for loan applications or credit card limits. The method is part of a banking strategy trying to reduce the customer churn in a context of a competitive retail banking market. We rely on the Q-learning algorithm and on a mean reverting diffusion process to address this topic. It leads ultimately to a fitted Q-iteration update and a model-free and off-policy setting. The diffusion process reflects the time series observed in retail banking such as transaction payments or credit card transactions. Such data is, however, strictly confidential and protected by the regulators, and therefore, it cannot be released publicly. We furthermore introduce a new terminal digital function, 
$\Pi$, defined as a Heaviside step function in its discrete form for a discrete variable $n \in \mathbb{R}$. The digital function is at the core of our approach for retail banking since it can evaluate the future probability of an event including, for instance, the future default probability of a client based on his spendings. Our method converges to an optimal policy, and to optimal sets of actions and states, respectively the spendings and the available money. The retail banks can, consequently, determine the optimal policy of money management based on the aggregated financial transactions of the clients. The banks are able to compare the difference between the MQLV's optimal policy and the individual policy of each client. It contributes to an unbiased decision making process while offering transparency to the client. Our main contributions are summarized below:

\begin{itemize}
\item A new RL framework called MQLV, Modified Q-Learning for Vasicek, extending the initial QLBS framework \cite{halperin2017qlbs}. MQLV uses the theoretical foundation of RL learning and Q-Learning to build a financial RL framework based on a mean reverting diffusion process, the Vasicek model  \cite{vasicek1977equilibrium}, to simulate data, in order to reach ultimately a model-free and off-policy RL setting.  \\
\item The definition of a digital function to estimate the future probability of an event. 
The aim is to widen the application perspectives of MQLV by using a characteristic terminal function that is usable for a decision making process in retail banking  such as the estimation of the default probability of a client.  \\
\item The first application of Q-learning to determine the clients' optimal policy of money management in retail banking. MQLV leverages the clients aggregated financial transactions to define the optimal policy of money management, targeting the risk estimation of bank loan applications or credit cards. 
\end{itemize}

The paper is structured as follows. We review QLBS and the Q-Learning formulations derived by Halperin in \cite{halperin2017qlbs} in the context of the Black, Scholes and Merton model in section \ref{sec::background}. We describe MQLV according to the Q-Learning algorithm that leads to a model-free and off-policy setting in section \ref{sec::algorithm}. 
We highlight experimental results in section \ref{sec::experiments}. We discuss related works in section \ref{sec::relatedwork} and we conclude in section \ref{sec::conclusion} by addressing promising directions for future work. 

\section{Background} \label{sec::background}
We define $A_t \in \mathcal{A}$ the action taken at time $t$ for a given state $X_t \in \mathcal{X}$ and the immediate reward by $R_{t+1}$. 
The ongoing state is denoted by $X_t \in \mathcal{X}$ and the stochastic diffusion process by $S_t \in \mathcal{S}$ at time $t$. The discount factor that trades off the importance of immediate and later rewards is expressed by $\gamma \in [0;1]$.  \\

We recall a policy is a mapping from states to probabilities of selecting each possible action \cite{sutton2018reinforcement}. By following the notations of \cite{halperin2017qlbs}, the policy $\pi$ such that 

\begin{equation} \label{eq::policy}
\pi : \left\lbrace 0, \ldots, T-1 \right\rbrace \times \mathcal{X}\rightarrow \mathcal{A}
\end{equation}

maps at time $t$ the current state $X_t=x_t$ into the action $a_t \in \mathcal{A}$.

\begin{equation} \label{eq::policymap}
a_t = \pi(t,x_t)
\end{equation}

The value of a state $x$ under a policy $\pi$, denoted by $v_\pi(x)$ when starting in $x$ and following $\pi$ thereafter, is called the state-value function for policy $\pi$.

\begin{equation} \label{eq::statevalue}
v_\pi = \mathbb{E}_\pi \left[ \sum_{k=0}^{\infty} \gamma^k R_{t+k+1} | X_t = x\right]
\end{equation}

The action-value function, $q_\pi (x,a)$ for policy $\pi$ defines the value of taking action $a$ in state $x$ under a policy $\pi$	as the expected return starting from $x$, taking the action $a$, and thereafter following policy $\pi$.

\begin{equation} \label{eq::actionvalue}
q_\pi(x,a) = \mathbb{E}_\pi \left[ \sum_{k=0}^{\infty} \gamma^k R_{t+k+1} | X_t = x, A_t = a \right]
\end{equation} 

The optimal policy, $\pi_t^*$, is the policy that maximizes the state-value function.

\begin{equation} \label{eq::optimalpolicy}
\pi_t^* (X_t) = \arg \max_\pi V_t^\pi (X_t)
\end{equation}

The optimal state-value function, $V_t^*$, satisfies the Bellman optimality equation such that

\begin{equation} \label{eq::optimalvalue}
V_t^* (X_t) = \mathbb{E}_t^{\pi^*} \left[ R_t(X_t, u_t=\pi_t^*(X_t), X_{t+1})+\gamma V_{t+1}^*(X_{t+1}) \right] .
\end{equation}

The Bellman equation for the action-value function, the Q-function, is defined as 

\begin{equation} \label{eq::bellmanQfunction}
Q_t^\pi (x,a) = \mathbb{E}_t \left[ R_t(X_t, a_t, X_{t+1}) | X_t = x, a_t = a\right] + \gamma \mathbb{E}_t^\pi \left[ V_{t+1}^\pi (X_{t+1}) | X_t = x \right] .
\end{equation}

The optimal action-value function, $Q_t^*$, is obtained for the optimal policy with 

\begin{equation} \label{eq::optimalQfunction}
\pi_t^* = \arg \max_\pi Q_t^\pi (x,a) .
\end{equation}

The optimal state-value and action-value functions are connected by the following system of equations.

\begin{equation} \label{eq::actstate}
\begin{cases}
 V_t^* =  \max_a Q^*(x,a) &  \\ 
 Q_t^* = \mathbb{E}_t \left[ R_t(X_t,a,X_{t+1}) \right] + \gamma \mathbb{E}_t \left[ V_{t+1}^* (X_{t+1} | X_t = x) \right] &  \\ 
\end{cases}
\end{equation} 

Therefore, we can obtain the Bellman optimality equation.
\begin{equation} \label{eq::bellmanoptimal}
Q_t^*(x,a) = \mathbb{E}_t \left[ R_t(X_t,a_t,X_{t+1}) + \gamma \max_{a_{t+1} \in \mathcal{A}} Q_{t+1}^* (X_{t+1}, a_{t+1}) | X_t=x, a_t=a \right]
\end{equation} 

Using the Robbins-Monro update \cite{robbins1985stochastic}, the update rule for the optimal Q-function with on-line Q-learning on the data point $(X_t^{(n)}, a_t^{(n)}, R_t^{(n)}, X_{t+1}^{(n)})$ is expressed by the following equation with $\alpha$ a constant step-size parameter.

\begin{equation} \label{eq::qlearning_update}
\begin{split}
Q_t^{*,k+1} (X_t, a_t) = & (1-\alpha^k) Q_t^{*,k}(X_t,a_t) + \\ 
& \: \alpha^k \left[ R_t(X_t, a_t, X_{t+1}) + \gamma \max_{a_{t+1} \in \mathcal{A}} Q_{t+1}^{*,k} (X_{t+1}, a_{t+1}) \right]
\end{split}
\end{equation}

\section{Algorithm} \label{sec::algorithm}
We describe, in this section, how to derive a general recursive formulation for the optimal action. It is equivalent to an optimal hedge under a financial framework such as, for instance, portfolio or personal finance optimization. We additionally present the formulation of the action-value function, the Q-function. Both the optimal hedge and the Q-function follow the assumption of a continuous space scenario generated by the Vasicek model with Monte Carlo simulation.  \\

By relying on the financial framework established in \cite{halperin2017qlbs}, we consider a mean reverting diffusion process, also known as the Vasicek model \cite{vasicek1977equilibrium}.

\begin{equation} \label{eq::meanreverting}
dS_t = \kappa(b-S_t)dt + \sigma dB_t
\end{equation}

The term $\kappa$ is the speed reversion, $b$ the long term mean level, $\sigma$ the volatility and $B_t$ the Brownian motion. The solution of the stochastic equation is equal to

\begin{equation} \label{eq::meanrevertingsol}
S_t = S_0 e^{-\kappa t} + b(1-e^{-\kappa t}) + \sigma e^{-\kappa t} \int_0^t e^{\kappa s}dB_s .
\end{equation}

Therefore, we define a new time-uniform state variable, i.e. without a drift, as 

\begin{equation} \label{eq::statevariable}
\begin{cases}
S_t = X_t + S_0e^{-\kappa t} + b(1-e^{-\kappa t}) & \\ 
\text{with } X_t = \sigma e^{-\kappa t} \int_0^t e^{\kappa s} dB_s - \left[ S_0e^{-\kappa t} + b(1-e^{-\kappa t}) \right] &
\end{cases} .
\end{equation}

Instead of estimating the price of a vanilla option as proposed in \cite{halperin2017qlbs}, we are interested to estimate the future probability of an event using the Q-learning algorithm and a digital function. First, we define the terminal condition reflecting that with the following equation

\begin{equation} \label{eq::terminalcondition}
Q_T^*(X_T, a_T=0) = -\Pi_T - \lambda Var \left[ \Pi_T(X_T) \right]
\end{equation} 

where $\Pi_T$ is the digital function at time $t=T$ defined such that

\begin{equation} \label{eq::digital}
\Pi_T = 1_{S_T\geq K}=
\left\{\begin{matrix}
& 1 \text{ if } S_T \geq K \\ 
& 0 \text{ otherwise}
\end{matrix}\right.
\end{equation}

and the second term, $\lambda Var \left[ \Pi_T(X_T) \right]$, is a regularization term with $\lambda \in \mathbb{R}^+ \ll 0$. We use a backward loop to determine the value of $\Pi_t$ for $t=T-1, ..., 0$.

\begin{equation} \label{eq::backward_pi}
\Pi_t = \gamma \left( \Pi_{t+1} - a_t \Delta S_t \right) \quad \text{with} \quad \Delta S_t = S_{t+1} -\frac{S_{t}}{\gamma} = S_{t+1} - e^{r \Delta t} S_t
\end{equation}

Following the definition of the equations (\ref{eq::optimalvalue}) and (\ref{eq::backward_pi}), we express the one-step time dependent random reward with respect to the cross-sectional information $\mathcal{F}_t$ as follows. 

\begin{equation} \label{eq::randomreward}
\begin{split}
R_t(X_t, a_t, X_{t+1}) & =  \gamma a_t \Delta S_t(X_t, X_{t+1}) - \lambda Var \left[ \Pi_t | \mathcal{F}_t \right]\\
& \text{with } Var \left[ \Pi_t | \mathcal{F}_t \right] = \gamma^2 \mathbb{E}_t \left[ \hat{\Pi}_{t+1}^2 -2a_t \Delta \hat{S}_t \hat{\Pi}_{t+1} + a_t^2 \Delta \hat{S}_t^2 \right]
\end{split}
\end{equation}

The term $\Delta \bar{S}_t$ is defined such that $\Delta \bar{S}_t = \frac{1}{N}\Delta S$,  $\Delta \widehat{S} = \Delta S - \Delta \bar{S}_t$ and $\hat{\Pi}_{t+1}=\Pi_{t+1} - \bar{\Pi}_{t+1}$ with $\bar{\Pi}_{t+1} = \frac{1}{N} \Pi_{t+1}$. Because of the regularizer term, the expected reward $R_t$ is quadratic in $a_t$ and has a finite solution. We therefore inject the one-step time dependent random reward equation (\ref{eq::randomreward}) into the Bellman optimality equation (\ref{eq::bellmanoptimal}) to obtain the following Q-learning update, $Q^\ast$, and the optimal action, $a^\ast$, to be solved within a backward loop $\forall t = T-1, ..., 0$.

\begin{equation} \label{eq::qlearningupdate}
\begin{split}
Q_t^\ast (X_t, a_t) = & \: \gamma \mathbb{E}_t \left[ Q_{t+1}^\ast (X_{t+1}, a_{t+1}^\ast ) + a_t \Delta S_t \right] - \lambda Var \left[ \Pi_t | \mathcal{F}_t \right]  \\ 
a_t^\ast (X_t) = & \: \mathbb{E}_t \left[ \Delta \hat{S}_t \hat{\Pi}_{t+1} + \frac{1}{2 \lambda \gamma} \Delta S_t \right] \left[\mathbb{E}_t \left[ \left( \Delta \hat{S}_t \right)^2 \right] \right]^{-1}
\end{split}
\end{equation}



We refer to \cite{halperin2017qlbs} for further details about the analytical solution, $a^\ast$, of the Q-learning update (\ref{eq::qlearningupdate}). Our approach uses the $N$ Monte Carlo paths simultaneously to determine the optimal action $a^*$ and the optimal action-value function $Q^*$ to learn the policy $\pi^\ast$. We thus do not need an explicit conditioning of $X_t$ at time $t$. We assume a set of basis function $\lbrace \Phi_n(x) \rbrace$ for which the optimal action $a_t^*(X_t)$ and the optimal action-value function, $Q_t^*(X_t,a_t^*)$, can be expanded. 

\begin{equation} \label{eq::basisfunction}
a_t^*(X_t) = \sum_n^M \phi_{nt} \Phi_n(X_t) \quad \text{and} \quad
Q_t^*(X_t, a_t^*) = \sum_n^M \omega_{nt} \Phi_n(X_t)
\end{equation}  

The coefficients $\phi$ and $\omega$ are computed recursively backward in time $\forall t = T-1, \ldots, 0$. We subsequently define the minimization problem to evaluate $\phi_{nt}$.

\begin{equation} \label{eq::minimization1}
G_t(\phi) = \sum_{k=1}^{N} \left[ -\sum_n^M \phi_{nt} \Phi_{n} (X_t^k) \Delta S_t^k + \gamma \lambda \left( \Pi_{t+1}^k - \sum_n^M \phi_{nt} \Phi_{n} (X_t^k) \Delta \widehat{S}_t^k\right)^2 \right]
\end{equation}

The equation (\ref{eq::minimization1}) leads to the following set of linear equations $\forall n = 1, \ldots, M$. 

\begin{equation} \label{eq::linearset1}
\begin{split}
\begin{dcases}
A_{nm}^{(t)} = \sum_{k=1}^N \Phi_n(X_t^k) \Phi_m(X_t^k) (\Delta \widehat{S}_{t^k})^2 & \\ 
B_n^{(t)} = \sum_{k=1}^N \Phi_n (X_t^k) \left[ \widehat{\Pi}_{t+1}^k \Delta \widehat{S}_t^k + \dfrac{1}{2\gamma \lambda} \Delta S_t^k \right] &
\end{dcases} 
\text{ with } \sum_m^M A_{nm}^{(t)} \phi_{mt} = B_n^{(t)}
\end{split}
\end{equation}

Therefore, the coefficients of the optimal action $a_t^*(X_t)$ are determined by 

\begin{equation} \label{eq::phi}
\phi_t^* = A_t^{-1} B_t .
\end{equation}

We hereinafter use the Fitted Q Iteration (FQI) \cite{hasselt2010double,murphy2005generalization} to evaluate the coefficients $\omega$. The optimal action-value function, $Q^*(X_t, a_t)$, is represented in its matrix form according to the basis function expansion of the equation (\ref{eq::basisfunction}).

\begin{equation} \label{eq::expansionQ}
\begin{split}
Q_t^*(X_t,a_t) = & \left(1, a, \dfrac{1}{2} a_t^2 \right) 
\begin{pmatrix}
W_{11}(t) & W_{12}(t) & \ldots & W_{1M}(t)  \\ 
W_{21}(t) & W_{22}(t) & \ldots & W_{2M}(t)  \\ 
W_{31}(t) & W_{32}(t) & \ldots & W_{3M}(t)
\end{pmatrix}
\begin{pmatrix}
\Phi_1 (X_t) \\ \vdots \\ \Phi_M (X_t)
\end{pmatrix} \\
= & A_t^T W_t \Phi(X_t) = A_t^T U_W(t,X_t)
\end{split}
\end{equation}


Based on the least-square optimization problem, the coefficients $W_t$ are determined using backpropagation $\forall t=T-1, ..., 0$ as follows 

\begin{equation} \label{eq::leastsquareW}
\begin{split}
\mathcal{L}_t(W_t) & = \sum_{k=1}^N \left( R_t(X_t,a_t,X_{t+1}) + \gamma \max_{a_{t+1}\in \mathcal{A}} Q_{t+1}^* (X_{t+1}, a_{t+1}) - W_t \Psi_t (X_t, a_t) \right)^2 \\
& \text{with } W_t \Psi (X_t, a_t)+\epsilon \underset{\epsilon \rightarrow 0}{\longrightarrow} R_t(X_t,a_t,X_{t+1}) + \gamma \max_{a_{t+1} \in \mathcal{A}} Q_{t+1}^* (X_{t+1}, a_{t+1})
\end{split}
\end{equation}

for which we derive the following set of linear equations.


\begin{equation} \label{eq::weightsW}
\begin{dcases}
M_n^{(t)} = \sum_{k=1}^{N} \Psi_n (X_t^k, a_t^k) \left[ \eta  \left( R_t(X_t,a_t,X_{t+1}) + \gamma \max_{a_{t+1} \in \mathcal{A}} Q_{t+1}^* (X_{t+1}, a_{t+1}) \right) \right]  & \\
\text{with } \eta \sim B(N, p)  &
\end{dcases}
\end{equation}

The term $B(N,p)$ represents the binomial distribution for $n$ samples with probability $p$. It plays the role of a dropout function when evaluating the matrix $M_t$ to compensate the well-known drawback of the Q-learning algorithm that is the overestimation of the Q-function values. We reach finally the definition of the optimal weights to determine the optimal action $a^\ast$. 

\begin{equation} \label{eq::optimalW}
W_t^* = S_t^{-1} M_t
\end{equation} 

The proposed model does not require any assumption on the dynamics of the time series, neither transition probabilities nor policy or reward functions. It is an off-policy model-free approach. The computation of the optimal policy, the optimal action and the optimal Q-function that leads to the future event probabilities is summed up in algorithm \ref{algo::mqlv}.

\SetAlFnt{\footnotesize} 
\SetAlCapFnt{\footnotesize} 
\SetAlCapNameFnt{\footnotesize} 
\SetKwFor{Case}{case}{}{}
\SetKwFunction{KwFn}{print}

\begin{algorithm}[t] 
\setstretch{1.25} 
\DontPrintSemicolon 
\KwData{time series of maturity T, either from generated or true data} 
\KwResult{optimal Q-function $Q^\ast$, optimal action $a^\ast$, value of digital function $\Pi$}

\Begin{ 

/*\textit{\small Condition at $T$}*/ 

$a_T^*(X_T) = 0 $

$Q_T(X_T, a_T) = - \Pi_T = -1_{S_T \geq K}$ using equation (\ref{eq::digital})

$Q_T^*(X_T, a_T^*) = Q_T(X_T, a_T)$

\vspace{.25cm}
/*\textit{\small Backward Loop}*/ 

\For{$t \gets T-1$ \KwTo $0$}{

    /*\textit{\small Evaluate the coefficients $\phi$}*/ 

    compute $A_t, B_t$ using equation (\ref{eq::linearset1})

    $\phi_t^* \gets A_t^{-1} B_t$

	/*\textit{\small Evaluate the coefficients $\omega$}*/ 

    compute $S_t, M_t$ using equation (\ref{eq::weightsW})

    $W_t^* \gets S_t^{-1} M_t$

    $a_t^*(X_t) = \sum_n^M \phi_{nt}^* \Phi_n(X_t)$

    $Q^*(X_t,a_t) = A_t^T W_t^* \Phi_(X_t)$
    }

\vspace{.25cm}
/*\textit{\small Compute the digital function value to estimate the event probability at $t=0$}*/ 

\KwFn{$\Pi_0 = \text{mean}(Q_0^*)$}

} 

\KwRet{} 

\caption{Q-learning to evaluate the optimal policy of money management \label{algo::mqlv}} 
\end{algorithm}

\section{Experiments} \label{sec::experiments}
We empirically evaluate the performance of MQLV. We initially highlight the similarities between historical payment transactions and Vasicek generated transactions. We then underline the MQLV's capabilities to learn the optimal policy of money management based on the estimation of future event probabilities in comparison to the closed formula of \cite{black1973pricing,merton1973theory}, hereinafter denoted by BSM's closed formula. We rely on synthetic data sets because of the privacy and the confidentiality issues of the retail banking data sets.
\\

\textbf{Data Availability and Data Description}
One of our contributions is to bring a RL framework designed for retail banking. However, none of the data sets can be released publicly because of the highly sensitive information they contain. We therefore show the similarities between a small sample of anonymized transactions and Vasicek generated transactions \cite{vasicek1977equilibrium}. We then use the Vasicek mean reverting stochastic diffusion process to generate larger synthetic data sets similar to the original retail banking data sets. The mean reverting dynamic is particularly interesting since it reflects a wide range of retail banking transactions including the credit card transactions, the savings history or the clients' spendings. Three different data sets were generated to avoid any bias that could have been introduced by using only one data set. We choose to differentiate the number of Monte Carlo paths between the data sets to assess the influence of the sampling size on the results. The first, second and third data sets contain respectively 20,000, 30,000 and 40,000 paths. We release publicly the data sets\footnote{\label{note1}The code and the data sets are available at https://github.com/dagrate/MQLV.} to ensure the reproducibility of the experiments.  \\

\textbf{Experimental Setup and Code Availability}
In our experiments, we generate synthetic data sets using the Vasicek model with a parameter $S_0=1.0$ corresponding to the value of the time series at $t=0$, a maturity of six months $T=0.5$, a speed reversion $a=0.01$, a long term mean $b=1$ and a volatility $\sigma=0.15$. The numbers were fixed such that any limitations of the methodology would be quickly observed because the choice of the parameters of the Vasicek model does not have any influence on the results of the Q-learning approach. The number of time steps is fixed equal to 5. We additionally use different strike values for the experiments explicitly mentioned in the Results and Discussions subsection. The simulations were performed on a computer with 16GB of RAM, Intel i7 CPU and a Tesla K80 GPU accelerator. To ensure the reproducibility of the experiments, the code is available at the following address\textsuperscript{\ref{note1}}. \\

\textbf{Results and Discussions about MQLV}
As aforementioned, we cannot release publicly an anonymized transactions data set because of privacy, confidentiality and regulatory issues. We consequently highlight the similarities between the dynamic of a small sample of anonymized transactions and Vasicek generated transactions for one client \cite{santandercreditcards} in figure \ref{fig::real_vs_gen}. The financial transactions in retail banking are periodic and often fluctuates around a long term mean, reflecting the frequency and the amounts of the spendings habits of the clients. 
The akin dynamic of the original and the generated transactions is highlighted by the small RMSE of 0.03. We also performed a least square calibration of the Vasicek parameters to assess the model's plausibility. We can observe in table \ref{tab::vas_vs_real} that the Vasicek parameters have the same magnitude and, therefore, it supports the hypothesis that the Vasicek model could be used to generate synthetic transactions.
\\

\begin{table}[t]
\begin{minipage}{0.48\textwidth}
 \frame{\includegraphics[scale=0.4]{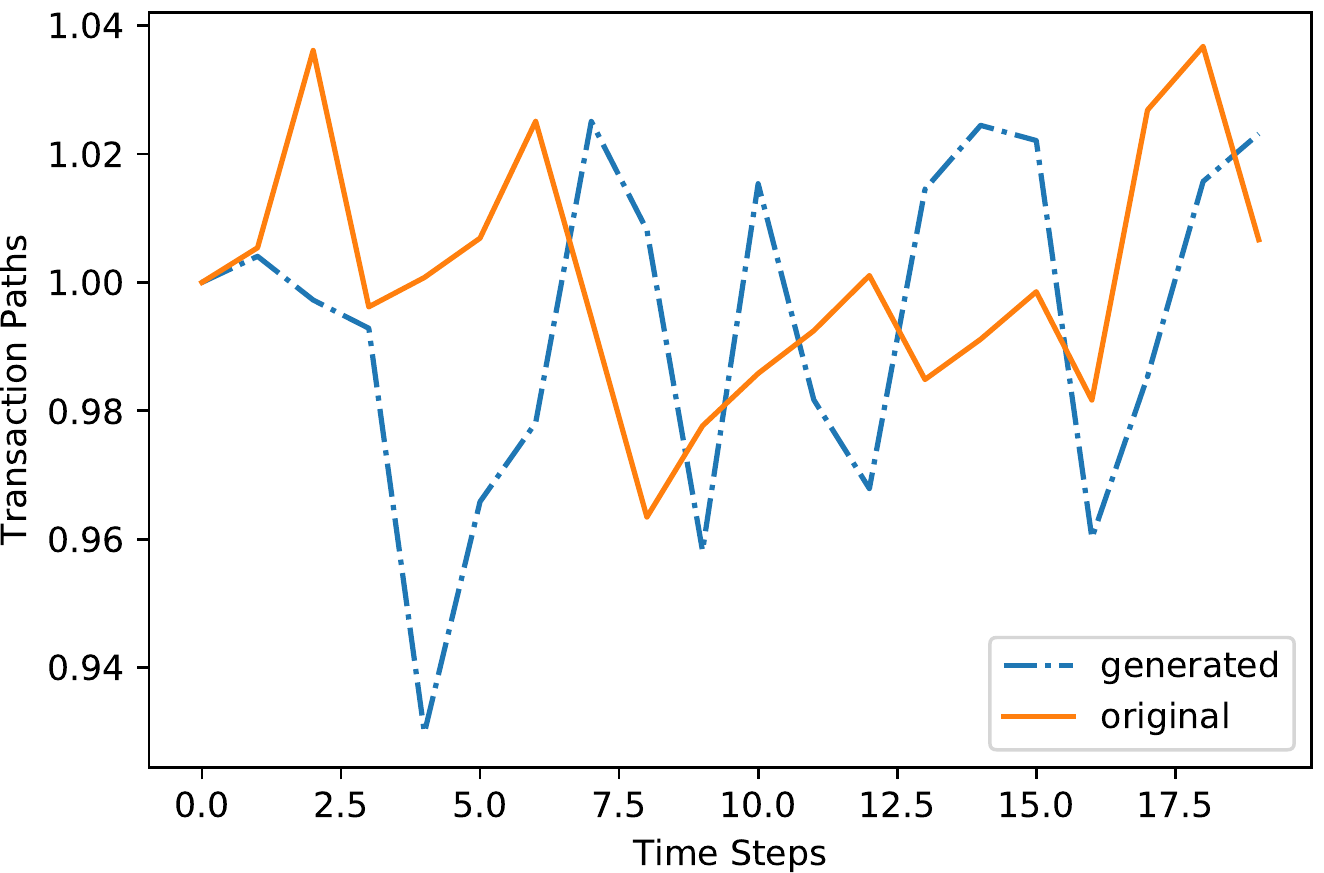}}
 \captionof{figure}{Samples of original and Vasicek generated transactions for one client. The two samples oscillate around a long term mean of 1 and have a similar pattern, highlighted by the small RMSE of 0.03 in table \ref{tab::vas_vs_real}.}
 \label{fig::real_vs_gen}
\end{minipage} \hfill
\begin{minipage}{0.48\textwidth}
\centering 
\caption{RMSE error between the samples of original transactions and generated Vasicek transactions of figure \ref{fig::real_vs_gen}. We also calibrated the Vasicek parameters according to the original transactions to validate the model's plausibility. } \label{tab::vas_vs_real}
\scalebox{0.9}{ 
\begin{tabular}{cccccc} 
  \toprule 
  \; Description \hspace{.25cm} & \hspace{.25cm} Value \;\\
  \midrule
  RMSE & 0.0335 \\
  Vasicek speed reversion $a$ & 0.5444 \\
  Vasicek long term mean $b$ & 0.9001 \\
  Vasicek volatility $\sigma$ & 0.2185 \\
  \bottomrule 
\end{tabular}
} 
\end{minipage}
\end{table}

\begin{figure}[!p]
 \centering
 \frame{\includegraphics[scale=0.65]{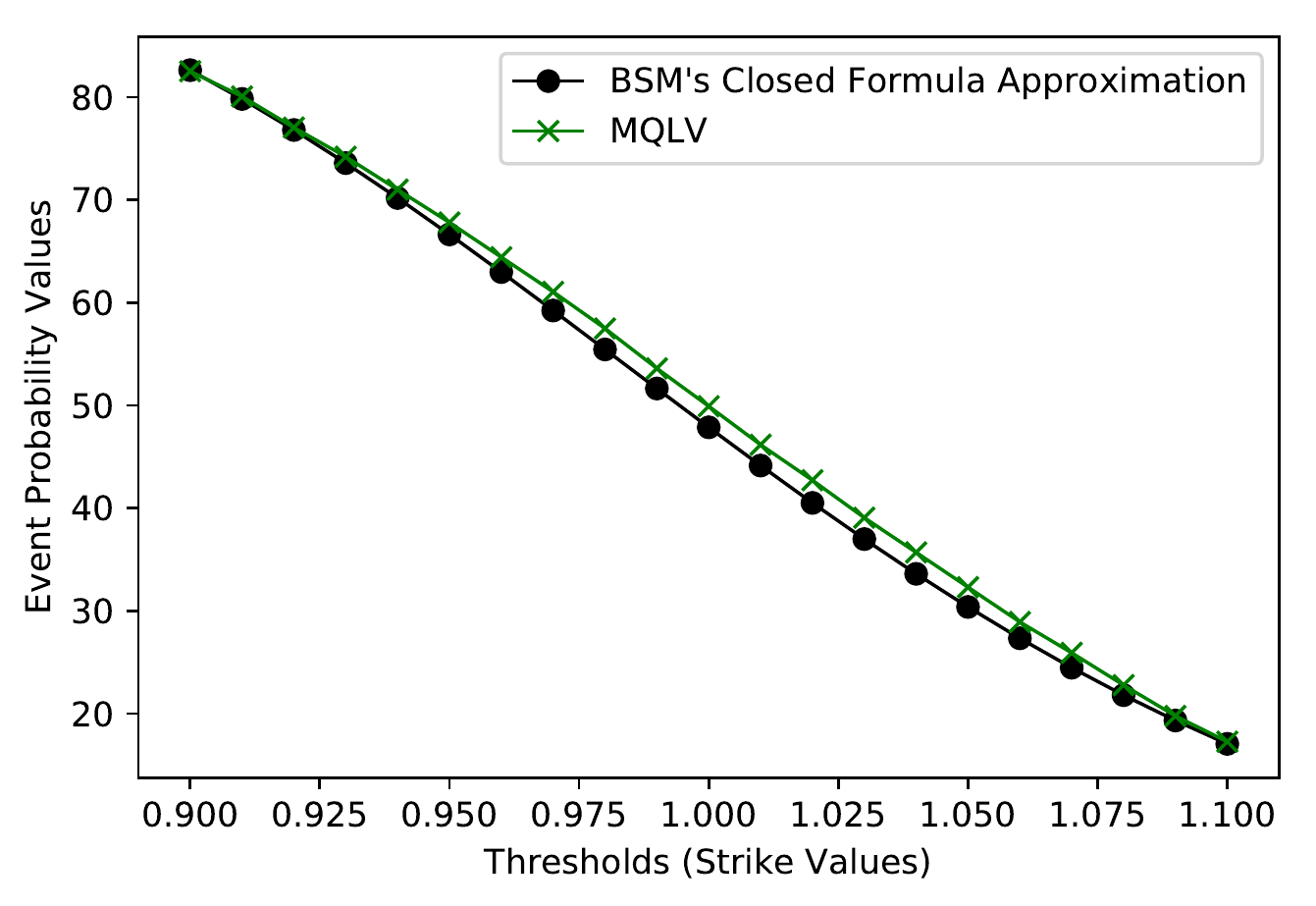}}
 \caption{Event probability values calculated by MQLV and BSM's closed formula approximation for 40,000 Monte Carlo paths with Vasicek parameters $a=0.01, b=1$ and $\sigma=0.15$. The BSM's closed formula approximation values are used as reference values. The event probabilities of MQLV are close to the BSM's values with a total RMSE of 1.502. It illustrates that MQLV is able to learn the optimal policy leading to accurate event probabilities.
 }
 \label{fig::digital}
\end{figure}

We present the core of our contribution in the following experiment. We aim at learning the optimal policy of money management. It is particularly interesting for bank loan applications where the differences between a client's spendings policy and the optimal policy can be compared. We show that MQLV is capable of evaluating accurately the probability of a default event using a digital function, which highlights the learning of the optimal policy of money management. Effectively, if the MQLV's learned policy is different than the optimal policy, then the probabilities of default events are not accurate. The estimation of future event probabilities for different strike values is represented in figure \ref{fig::digital}. We rely on the BSM's closed formula for the vanilla option pricing \cite{black1973pricing,merton1973theory} to approximate the digital function values \cite{hull2003options}. We used, therefore, the BSM's values as reference values to cross-validate the MQLV's values. MQLV achieves a close representation of the event probabilities for the different strike values in figure \ref{fig::digital}. The curves of both the MQLV and the BSM's approaches are similar with a RMSE of 1.5016. This result highlights that the learned Q-learning policy of MQLV is sufficiently close to the optimal policy to compute event probabilities almost identical to the probabilities of the BSM's formula approximation.  \\

We gathered quantitative results in table \ref{tab::digital} for a deeper analysis of the MQLV's results. The event probability values are listed for the three data sets. We chose a set of parameters for the Vasicek model such that our configuration is free of any time-dependency. We therefore expect a probability value of 50\% at a threshold of 1 because the standard deviation of the generated data sets is only induced by the standard deviation of the normal distribution, used to simulate the Brownian motion. Surprisingly, 
the MQLV values at a strike of 1 are closer to 50\% than the BSM's values for all the data sets. We can conclude, subsequently, that, for our configuration, MQLV is capable to learn the optimal policy of money management which is reflected by the accurate evaluation of the event probabilities. \\

\begin{table}[!p] 
\centering 
\caption{Valuation differences of the digital values for event probabilities according to different strikes between the BSM's closed formula approximation and MQLV. Given our time-uniform configuration, the event probability values should be close to 50\% for a strike value of 1. The MQLV values are close to the theoretical target of 50\% at a strike of 1 highlighting the MQLV's capabilities to learn the optimal policy. The BSM's closed formula approximation slightly underestimates the probability values.} \label{tab::digital}
\scalebox{1.0}{ 
\begin{tabular}{cccccc}
  \toprule 
  \; Data \; & \; Number \; & \; Strike \; & \; BSM's Approx. \; & \; MQLV \; & \; Absolute \\ 
  \; Set \; & \; of Paths \; & \; Values \; & \; Values (\%) \; & \; Values (\%) \; & \; Difference \\ 
  \midrule
  1 & 20,000 & 0.92 & 76.810 & \textbf{77.098} & 0.288 \\
  1 & 20,000 & 0.98 & 55.447 & \textbf{57.920} & 2.473 \\
  1 & 20,000 & 1.00 & 47.867 & \textbf{50.235} & 2.368 \\
  1 & 20,000 & 1.02 & 40.509 & \textbf{42.865} & 2.356 \\
  2 & 30,000 & 0.92 & 76.810 & \textbf{76.953} & 0.143 \\
  2 & 30,000 & 0.98 & 55.447 & \textbf{57.760} & 2.313 \\
  2 & 30,000 & 1.00 & 47.867 & \textbf{50.043} & 2.176 \\
  2 & 30,000 & 1.02 & 40.509 & \textbf{42.744} & 2.235 \\
  3 & 40,000 & 0.92 & 76.810 & \textbf{77.047} & 0.237 \\
  3 & 40,000 & 0.98 & 55.447 & \textbf{57.491} & 2.044 \\
  3 & 40,000 & 1.00 & 47.867 & \textbf{49.924} & 2.057 \\
  3 & 40,000 & 1.02 & 40.509 & \textbf{42.713} & 2.204 \\
  \bottomrule 
\end{tabular} 
} 
\end{table} 

\begin{table}[t] 
\centering 
\caption{Event probabilities for data sets generated with different Vasicek parameters $a$ and $\sigma$. The parameter $b$ remains unchanged to keep a configuration free of any time-dependency to facilitate the results explainability. 
We can deduce that MQLV is able to learn the optimal policy because the MQLV's probabilities are close to the theoretical target of 50\% at a strike of 1. MQLV is also more accurate than BSM's formula in this configuration. } \label{tab::otherdata}
\scalebox{0.975}{ 
\begin{tabular}{cccccc}
  \toprule 
  \; Parameters \; & \; Number \; & \; Strike \; & \; BSM's App. \; & \; MQLV \; & \; Absolute \\ 
  \; $a; b; \sigma$ \; & \; of Paths \; & \; Values \; & \; Values (\%) \; & \; Values (\%) \; & \; Difference \\ 
  \midrule
  0.01; 1; 0.10 & 50,000 & 0.98 & 59.856 & \textbf{61.223} & 1.366 \\
  0.01; 1; 0.10 & 50,000 & 1.00 & 48.562 & \textbf{50.001} & 1.439 \\
  0.01; 1; 0.10 & 50,000 & 1.02 & 37.596 & \textbf{39.044} & 1.447 \\
  0.01; 1; 0.30 & 50,000 & 0.98 & 49.558 & \textbf{53.647} & 4.089 \\
  0.01; 1; 0.30 & 50,000 & 1.00 & 45.767 & \textbf{49.997} & 4.230 \\
  0.01; 1; 0.30 & 50,000 & 1.02 & 42.088 & \textbf{46.194} & 4.106 \\
  0.10; 1; 0.15 & 50,000 & 0.98 & 55.447 & \textbf{57.540} & 2.093 \\
  0.10; 1; 0.15 & 50,000 & 1.00 & 47.867 & \textbf{50.015} & 2.148 \\
  0.10; 1; 0.15 & 50,000 & 1.02 & 40.509 & \textbf{42.638} & 2.129 \\
  0.30; 1; 0.15 & 50,000 & 0.98 & 55.447 & \textbf{57.586} & 2.139 \\
  0.30; 1; 0.15 & 50,000 & 1.00 & 47.867 & \textbf{50.022} & 2.155 \\
  0.30; 1; 0.15 & 50,000 & 1.02 & 40.509 & \textbf{42.542} & 2.033 \\
  \bottomrule 
\end{tabular} 
} 
\end{table} 

We chose to generate three new data sets 
with new Vasicek parameters $a$ and $\sigma$ to underline the potential of MQLV and the universality of the results. In table \ref{tab::otherdata}, we computed the event probabilities for different strikes 
for the newly generated data sets. The parameter $b$ remains unchanged since we want to keep a configuration free of any time-dependency. We notice that MQLV is capable to estimate a probability of 50\% for a strike of 1 which can only be obtained if MQLV is able to learn the optimal policy. We also observe that the BSM's approximation does lead to a lower accuracy. We showed in this experiment that our model-free and off-policy RL approach, MQLV, is able to learn the optimal policy reflected by the accurate probability values independently of the data sets considered and of the Vasicek parameters.  \\

\textbf{Limitations of the BSM's closed formula used for cross validation} In our experiments, we observed, surprisingly,  that the BSM's closed formula approximation underestimates the event probability values. The volatility is the only parameter playing a significant role in the generation of the time series and, therefore, the event probability should be equal to the mean of the distribution used to generate the random numbers. The Brownian motion is simulated with a standard normal distribution with a 0.5 mean. The BSM's closed formula did not, however, lead to a probability of 0.5 but to slightly smaller values because of the limit of their theoretical framework \cite{black1973pricing,merton1973theory}. We hence observed that MQLV was more accurate than the BSM's closed formula in our configuration.

\section{Related Work} \label{sec::relatedwork}
The foundations of modern reinforcement learning described in \cite{sutton1984temporal,williams1987class} established the theoretical framework to learn good policies for sequential decision problems by proposing a formulation of cumulative future reward signal. The Q-learning algorithm introduced in \cite{watkins1989learning} is one of the cornerstone of all recent reinforcement learning publications. However, the convergence of the Q-Learning algorithm was solved several years later. It was shown that the Q-Learning algorithm with non-linear function approximators \cite{tsitsiklis1997analysis} with off-policy learning \cite{baird1995residual} could provoke a divergence of the Q-network. The reinforcement learning community therefore focused on linear function approximators \cite{tsitsiklis1997analysis} to ensure convergence.  \\

The emergence of neural networks and deep learning \cite{goodfellow2016deep} contributed to address the use of reinforcement learning with neural networks. At an early stage, deep auto-encoders were used to extract feature spaces to solve reinforcement learning tasks \cite{lange2010deep}. Thanks to the release of the Atari 2600 emulator \cite{bellemare2013arcade}, a public data set then was available answering the needs of the RL community for larger simulation. The Atari emulator allowed a proper performance benchmark of the different reinforcement learning algorithms and offered the possibility to test various architectures. The Atari games were used to introduce the concept of deep reinforcement learning \cite{mnih2013playing,mnih2015human}. The authors used a convolutional neural network trained with a variant of Q-learning to successfully learn control policies directly from high dimensional sensory inputs. They reached human-level performance on many of the Atari games. Shortly after, the deep reinforcement learning was challenged by double Q-Learning within a deep reinforcement learning framework \cite{van2016deep}. The double Q-Learning algorithm was initially introduced in \cite{hasselt2010double} in a tabular setting. The double deep Q-Learning gave more accurate estimates and lead to much higher scores than the one observed in \cite{mnih2013playing,mnih2015human}. An ongoing work is consequently to further improve the results of the double deep Q-learning algorithms through different variants. The authors used a quantile regression to approximate the full quantile function for the state-action return distribution in \cite{dabney2018implicit}, leading to a large class of risk-sensitive policies. It allowed them to further improve the scores on the Atari 2600 games simulator. Similarly, a new algorithm, called C51, which applies the Bellman's equation to the learning of the approximate value distribution was designed in \cite{bellemare2017distributional}. They showed state-of-the-art results on the Atari 2600 emulator.  \\

Other publications meanwhile focused on model-free policies and actor-critic framework. Stochastic policies were trained in \cite{wawrzynski2013autonomous} with a replay buffer to avoid divergence. It was showed in \cite{silver2014deterministic} that deterministic policy gradients (DPG) exist, even in a model-free environment. The DPG approach was subsequently extended in \cite{balduzzi2015compatible} using a deviator network. Continuous control policies were learned using backpropagation introducing the Stochastic Value Gradient SVG(0) and SVG(1) in \cite{heess2015learning}. Recently, Deep Deterministic Policy Gradient (DDPG) was presented in \cite{lillicrap2015continuous} to learn competitive policies using an actor-critic model-free algorithm based on the DPG that operates over continuous action spaces.  \\

\section{Conclusion} \label{sec::conclusion}
We introduced Modified Q-Learning for Vasicek or MQLV, a new model-free and off-policy reinforcement learning approach capable of evaluating an optimal policy of money management based on the aggregated transactions of the clients. MQLV is part of a banking strategy that looks to minimize the customer churn by including more transparency and more personalization in the decision process related to bank loan applications or credit card limits. It relies on a digital function, a Heaviside step function expressed in its discrete form, to estimate the future probability of an event such as a payment default. We discuss its relation with the Bellman optimality equation and the Q-learning update. We conducted experiments on synthetic data sets because of the privacy and confidentiality issues related to the retail banking data sets. The generated data sets followed a mean reverting stochastic diffusion process, the Vasicek model, simulating retail banking data sets such as transaction payments. Our experiments showed the performance of MQLV with respect to the BSM's closed formula for vanilla options. We also highlighted that MQLV is able to determine an optimal policy, an optimal Q-function, the optimal actions and the optimal states reflected by accurate probabilities. Surprisingly, we observed that MQLV led to more accurate event probabilities than the popular BSM's formula in our configuration.

Future work will address the creation of a fully anonymized data set illustrating the retail banking daily transactions with a privacy, confidentiality and regulatory compliance. We will also evaluate the MQLV's performance for data sets that violate the Vasicek assumptions. 
We furthermore observed that the Q-learning update could minor the real probability values for simulation involving a small temporal discretization such as $\Delta t = 200$. Preliminary results showed it is provoked by the basis function approximator error. We will address this point in future research. We will finally extend the Q-learning update to other scheme for improved accuracy and incorporate a deep learning framework.
\bibliographystyle{./splncs04}
\bibliography{./zzz-mybibliography}
\end{document}